\documentclass[letterpaper, 10 pt, conference]{ieeeconf}  

\IEEEoverridecommandlockouts                              

\usepackage{cite}
\usepackage{amsmath,amssymb,amsfonts}
\usepackage{algorithmic}
\usepackage{graphicx}
\usepackage{textcomp}
\usepackage{xcolor}
\usepackage{mathtools}
\usepackage{booktabs}
\usepackage{siunitx}
\usepackage[export]{adjustbox}
\usepackage{balance}

\begin{document}

\title{\LARGE \bf
Controlling Steering Angle for Cooperative Self-driving Vehicles utilizing CNN and LSTM-based Deep Networks
}


\author{Rodolfo Valiente$^*$, Mahdi Zaman$^*$, Sedat Ozer$^\dagger$, Yaser P. Fallah$^*$\\
$^*$Center for Research in Electric Autonomous Transport (CREAT), Orlando, FL\\
$^\dagger$University of Central Florida, Orlando, FL\\ 
\{rvalienter90, mahdizaman\}@knights.ucf.edu, sedatist@gmail.com, yaser.fallah@ucf.edu\\
}

\maketitle
\begin{abstract}
A fundamental challenge in autonomous vehicles is adjusting the steering angle at different road conditions. Recent state-of-the-art solutions addressing this challenge include deep learning techniques as they provide end-to-end solution to predict steering angles directly from the raw input images with higher accuracy. Most of these works ignore the temporal dependencies between the image frames. In this paper, we tackle the problem of utilizing multiple sets of images shared between two autonomous vehicles to improve the accuracy of controlling the steering angle by considering the temporal dependencies between the image frames. This problem has not been studied in the literature widely. We present and study a new deep architecture to predict the steering angle automatically by using Long-Short-Term-Memory (LSTM) in our deep architecture. Our deep architecture is an end-to-end network that utilizes CNN, LSTM and fully connected (FC) layers and it uses both present and futures images (shared by a vehicle ahead via Vehicle-to-Vehicle (V2V) communication) as input to control the steering angle. Our model demonstrates the lowest error when compared to the other existing approaches in the literature.
\end{abstract}


\section{Introduction}

Controlling the steering angle is a fundamental problem for autonomous vehicles \cite{Bojarski2016}, \cite{Chen2017}, \cite{Eraqi2017}. Recent computer vision-based approaches to control the steering angle in autonomous cars mostly focus on improving the driving accuracy with the local data collected from the sensors on the same vehicle and as such, they consider each car as an isolated unit gathering and processing information locally.  However, as the availability and the utilization of V2V communication increases, real-time data sharing becomes more feasible among vehicles \cite{hnmahjoub:syscon}, \cite{hnmahjoub:vtc}, \cite{hnmahjoub:cavs}. As such, new algorithms and approaches are needed that can utilize the potential of cooperative environments to improve the accuracy and the effectiveness of self-driving systems. 

\par
\begin{figure}[t]
      \centering
   
      \includegraphics[width=.48\textwidth]{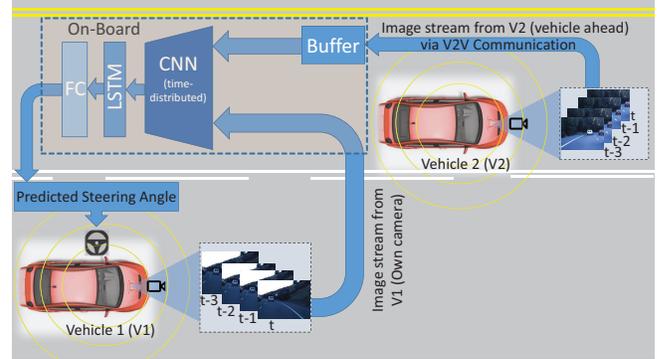}
      \caption{The overview of our proposed vehicle-assisted end-to-end system. Vehicle 2 (V2) sends his information to Vehicle 1 (V1) over V2V communication. V1 combines that information along with its own information to control the steering angle. The prediction is made through our CNN+LSTM+FC network (see Fig. \ref{fig:model} for the details of our network).}
      \label{fig:proposed_system}
  \end{figure}

In this paper, we present a deep learning-based approach that utilizes two sets of images coming from both the onboard sensors; e.g cameras; and from another vehicle ahead over V2V communication for the control of the steering angle in self-driving vehicles (see Fig. \ref{fig:proposed_system}). Our proposed deep architecture contains a convolutional neural network (CNN) followed by a LSTM and a FC network. Unlike the traditional approach, that manually decomposes the autonomous driving problem into different components as in \cite{Aly2008}, \cite{Alvarez2012} the end-to-end model can directly steer the vehicle from the camera data and has been proven to operate more effectively in previous works \cite{Bojarski2016}, \cite{Xu2017}. We compare our proposed deep architecture to multiple existing algorithms in the literature on Udacity dataset. Our experimental results demonstrate that our proposed CNN-LSTM-based model yields the state-of-the-art results. Our main contributions are: (1) we propose an end-to-end vehicle-assisted steering angle control system for cooperative systems; (2) We propose using a large sequence of images as opposed to using only two consecutive frames; (3) introduce a new deep architecture that obtain the state-of-the-art results on the Udacity dataset.

\section{Related Work}

The problem of navigating self-driving car by utilizing the perception acquired from sensory data has been studied in the literature with and without using end-to-end approaches \cite{schwarting2018planning}. For example the works from \cite{Agarwal2018}, \cite{Shin2018} use multiple components for recognizing objects of safe-driving concerns, such as lanes, vehicles, traffic signs and pedestrians. The recognition results are then combined to give a reliable world representation, which are used with an Artificial Intelligence (AI) system to make decisions and control the car.

Recent works focus on using end-to-end approaches\cite{amini2018variational}. The Autonomous Land Vehicle in a Neural Network (ALVINN) system was one of the earlier systems utilizing multilayer perceptron (MLP) \cite{Pomerleau1989} in 1989. Recently, CNNs were commonly used as in the DAVE-2 Project \cite{Bojarski2016}. In \cite{Eraqi2017}, the authors proposed an end-to-end trainable C-LSTM network that uses a LSTM network at the end of the CNN network. Similar approach was taken by the authors in \cite{Du2017}, who designed a 3D CNN model with residual connections and LSTM layers. Other researchers have implemented different variants of convolutional architecture for end-to-end models in \cite{Gurghian2016}, \cite{Dirdal2018}, \cite{Yu2017}. Another widely used approach for controlling vehicle steering angle in autonomous systems is via sensor fusion where combining image data with other sensor data such as LiDAR, RADAR, GPS improves the accuracy in autonomous operation \cite{Cho2014}, \cite{Gohring2011}. As an instance, in \cite{Dirdal2018}, the authors  designed a fusion network using both image features and LiDAR features based on VGGNet. 

All the above-listed work focus on utilizing the image data obtained from the on-board sensors and they do not consider the assisted data that comes from another car. In this paper, we demonstrate that using additional data that comes from the ahead vehicle helps us obtain better accuracy in controlling steering angle. In our approach, we utilize the information that is available to a vehicle ahead of our car to control the steering angle. The rest of the paper is organized as follows: In Section III the proposed approach is explained, Section IV provides details about the performed experiments. Finally, in Section V we conclude the paper and discuss possible directions for future work.

\section{Proposed Approach}

Controlling steering angle directly from input images is a regression-value problem. For that purpose, we can either use a single image or a sequence of (multiple) images. Considering multiple frames in a sequence can benefit us in situations where the present image alone is affected by noise or contains less useful information. For example, when the current image is burnt largely by direct sunlight or when the vehicle reaches a dead-end. In such situations, the correlation between the current frame and the past frames can be useful to decide the next steering value. To utilize multiple images as a sequence, we use LSTM. LSTM has a recursive structure acting as a memory, through which a network can keep some past information and solve for a regression value based on the dependency of the consecutive frames \cite{Gers2000}, \cite{Greff2017}.

Our proposed idea in this paper relies on the fact that the condition of the road ahead has already been seen by another vehicle recently and we can utilize that information to control the steering angle of our car as discussed above. Fig. \ref{fig:proposed_system} illustrates our approach. In the figure, Vehicle 1 receives a set of images from Vehicle 2 over V2V communication and keeps the data on board. It combines the received data with the data obtained from the onboard camera and processes those two sets of images on board to control the steering angle via an end-to-end deep architecture. This method enables the vehicle to look ahead of its current position at any given time.

\begin{figure}[t]
  \centering
  \includegraphics[width=.48\textwidth]{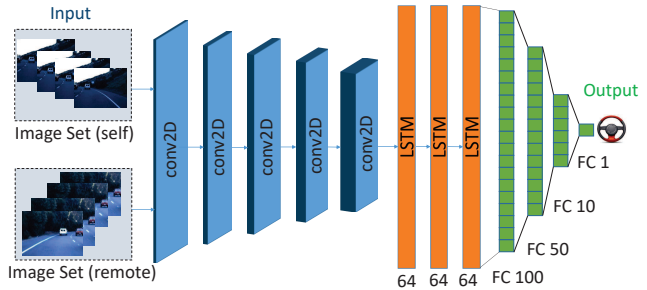}
  \caption{CNN + LSTM + FC Image sharing model. Our model uses 5 convolutional layers, followed by 3 LSTM layers, followed by 4 FC layers. See Table \ref{table:proposed_arq} for further details of our proposed architecture.}
  \label{fig:model}
\end{figure}

Our deep architecture is presented in Fig. \ref{fig:model}. The network takes the set of images from both vehicles as input and at the last layer, it predicts the steering angle as the regression output. The details of our deep architecture are given in Table \ref{table:proposed_arq}. Since we construct this problem as a regression problem with a single unit at the end, we use the Mean Squared Error (MSE) loss function in our network during the training. 

\begin{table}[t]
\caption{Details Of Proposed Architecture}
\begin{center}
 \begin{tabular}{c c c c c}
    Layer & Type & Size & Stride & Activation \\ [0.5ex]
 \hline\hline
 0 & Input & 640*480*3*2X & - & - \\ 
1 & Conv2D & 5*5, 24 Filters & (5,4) & ReLU \\
2 & Conv2D & 5*5, 32 Filters & (3,2) & ReLU \\
3 & Conv2D & 5*5, 48 Filters & (5,4) & ReLU \\
4 & Conv2D & 5*5, 64 Filters & (1,1) & ReLU \\
5 & Conv2D & 5*5, 128 Filters & (1,2) & ReLU \\
6 & LSTM & 64 Units & - & Tanh \\
7 & LSTM & 64 Units & - & Tanh \\
8 & LSTM & 64 Units & - & Tanh \\
9 & FC & 100  & - & ReLU \\
10 & FC & 50  & - & ReLU \\
11 & FC & 10  & - & ReLU \\
12 & FC & 1  & - & Linear \\
 \hline
 \hline
 \end{tabular}
\end{center}
\label{table:proposed_arq}
\end{table}

\section{Experiment Setup}
In this section we will elaborate further on the dataset as well as data preprocessing and evaluation metrics. We conclude the section with details of our implementation.
\subsection{Dataset}

\begin{figure}[t]
  \centering
  
   \includegraphics[width=.48\textwidth]{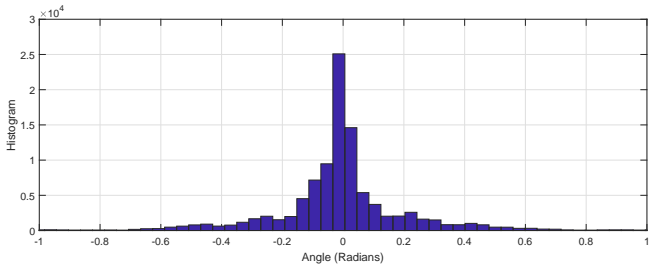}
  \caption{The angle distribution within the entire Udacity dataset (angle in radians vs. total number of frames), just angles between -1 and 1 radians are shown.}
  \label{fig:angle_distrib}
\end{figure}

In order to compare our results to existing work in the literature, we used the self-driving car dataset by Udacity. The dataset has a wide variation of 100K images from simultaneous Center, Left and Right camera on a vehicle, collected in sunny and overcast weather, 33K images belong to center camera. The dataset contains the data of 5 different trips with a total drive time of 1694 seconds. Test vehicle has 3 cameras mounted as in \cite{Bojarski2016} collecting images at a rate of near 20Hz. Steering wheel angle, acceleration, brake, GPS data was also recorded. The distribution of the steering wheel angles over the entire dataset is shown in Fig. \ref{fig:angle_distrib}. As shown in Fig. \ref{fig:angle_distrib}, the dataset distribution includes a wide range of steering angles. The image size is 480*640*3 pixels and total dataset is of 3.63 GB. Since there is no dataset available with V2V communication images currently, here we simulate the environment by creating a virtual vehicle that is moving ahead of the autonomous vehicle and sharing camera images by using the Udacity dataset.

\par
Udacity dataset has been used widely in the recent relevant literature \cite{Du2017},\cite{Choudhary2017} and we also use Udacity dataset in this paper to compare our results to the existing techniques in literature. Along with the steering angle, the dataset contains spatial (latitude, longitude, altitude) and dynamic (angle, torque, speed) information labelled with each image. The data format for each image is: index, timestamp, width, height, frame\_id, filename, angle, torque, speed, latitude, longitude, altitude. For our purpose, we are only using the sequence of center-camera images.

\subsection{Data Preprocessing}
The images in the dataset are recorded at the rate around 20 frame per second. Therefore, usually there is a large overlap between consecutive frames. To avoid overfitting, we used image augmentation to get more variance in our image dataset. Our image augmentation technic randomly adds brightness and contrast to change pixel values. We also tested image cropping to exclude possible redundant information that are not relevant in our application. However, in our test the models perform better without cropping.
\par
For the sequential model implementation, we preprocessed the data in a different way. Since we do want to keep the visual sequential relevance in the series of frames while avoiding overfitting, we shuffle the dataset while keeping track of the sequential information. We then train our model with 80\% images on the same sequence from the subsets and validate on the rest 20\%. 

\subsection{Vehicle-assisted Image Sharing}
Modern wireless technology allows us to share data between vehicles at high bitrates of up to Gbits/s (e.g., in peer-to-peer and line-of-sight mmWave technologies \cite{btoghi:vnc, btoghi:vtc2019}). Such communication links can be utilized to share images between vehicles for improved control. In our experiments, we simulate that situation between two vehicles as follows: we assume that both vehicles are away from each other by $\Delta t$ seconds. We take the $x$ consecutive frames ($t, t-1, ..., t-x+1$) from the self-driving vehicle (vehicle 1) at time step $t$ and the set of images containing $x$ future frames starting at ($t+ \Delta t$) from the other vehicle. Thus, a single input data (sample) contains a set of $2x$ frames for the model.

\subsection{Evaluation Metrics}
The steering angle is a continuous variable predicted for each time step over the sequential data and the metrics: mean absolute error (MAE) and root mean squared error (RMSE) are two of the most common used metrics in the literature to measure the effectiveness of the controlling systems. For example, RMSE is used in \cite{Du2017}, \cite{Choudhary2017} and MAE in \cite{Islam2018}. Both MAE and RMSE express average model prediction error and their values can range from 0 to $\infty $. They both are indifferent to the error sign. Lower values are better for both metrics. 
%

\subsection{Baseline Networks}

\begin{figure}[b]
  \centering
  \includegraphics[width=.48\textwidth]{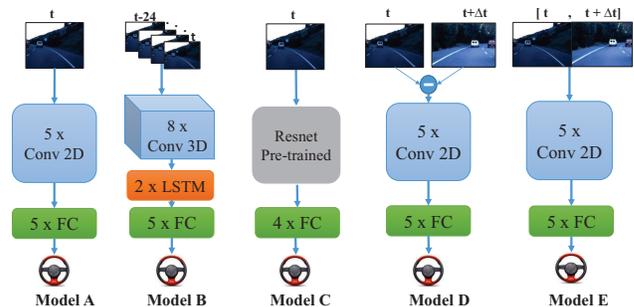}
  \caption{An overview of the used baseline models in this paper. The details of each model can be found in their respective source paper.}
  \label{fig:baseline_networks}
\end{figure}
As baseline, we include multiple deep architectures that have been proposed in the literature to compare our proposed algorithm. Those models from \cite{Bojarski2016}, \cite{Du2017} and \cite{Choudhary2017} are, to the best of our knowledge, the best reported approaches in the literature using a camera only. In total, we chose 5 baseline end-to-end algorithms to compare our results. We name these five models as models A, B, C, D and E in the rest of this paper. Model A is our implementation of the model presented in \cite{Bojarski2016}. Models B and C are the proposal of \cite{Du2017} . Models D and E are reproduced as in \cite{Choudhary2017}. The overview of these models is given in Fig. \ref{fig:baseline_networks}. Model A uses a CNN-based network while Model B combines LSTM with 3D-CNN and uses 25 time-steps as input. Model C is based on ResNet \cite{Wu2015} model and Model D uses the difference image of two given time-steps as input to a CNN-based network. Finally, Model E uses the concatenation of two images coming from different time-steps as input to a CNN-based network.

\subsection{Implementation and Hyperparameter Tuning}

Our implementations use Keras with a Tensor Flow backend. Final training is done on two NVIDIA Tesla V100 16GB GPUs. When implemented on our system, the training took 4 hours for the model in \cite{Bojarski2016} and between 9-12 hours for the deeper networks used in \cite{Du2017}, in \cite{Choudhary2017} and our proposed network. 
\par
We used Adam optimizer \cite{Hannah2015} in all our experiments (learning rate of $10^{-2}$, $\beta1 = 0.900$ , $\beta2  = 0.999$, $\mathcal{E} = 10^{-8}$). For learning rate, we tested from $10^{-1}$ to $10^{-6}$ and we found the best-performing learning rate being $10^{-3}$. We also studied the minibatch size to see its effect on our network. Minibatch sizes of 128, 64 and 32 are tested and the value 64 yielded the best results for us therefore we used 64 in our experiments reported in this paper.
\par
\begin{figure}[b]
  \centering
  \includegraphics[width=.48\textwidth]{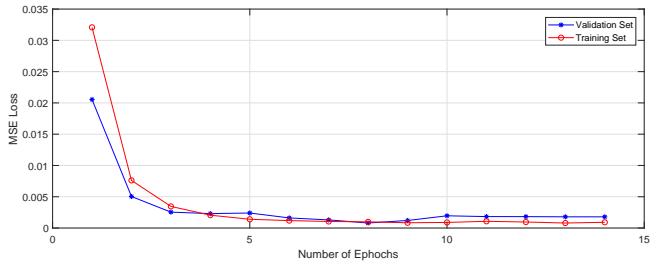}
  \caption{Training and Validation steps for our best model with $x=8$.}
  \label{fig:trainingVSvalidationx8}
\end{figure}

 
Fig. \ref{fig:trainingVSvalidationx8} demonstrates how the value of the loss function changes as the number of epochs increases for both training and validation data sets. The MSE loss decreases after the first few epochs rapidly and then remains stable, remaining almost constant around the 14\textsuperscript{th} epoch.

\begin{figure}[t]
  \centering
 \includegraphics[width=.48\textwidth]{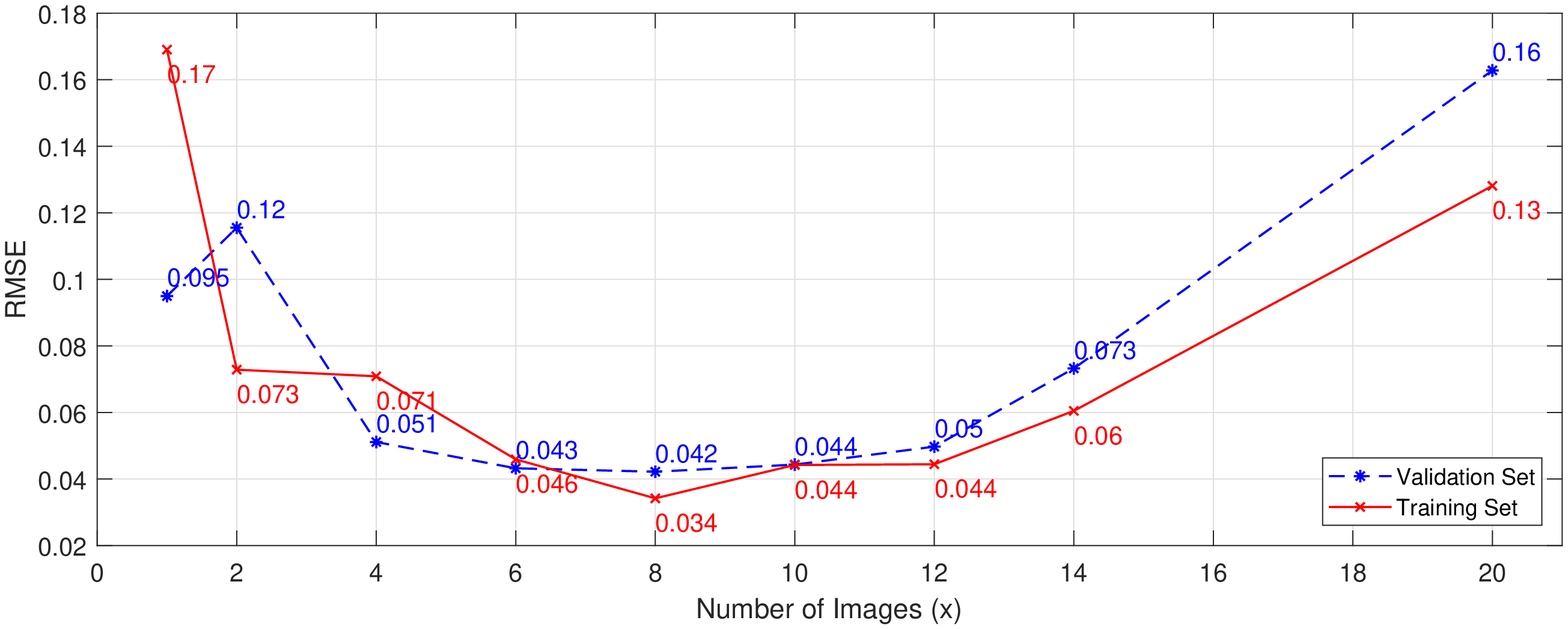}
  \caption{RMSE vs.$x$ value. We trained our algorithm at various $x$ values and computed the respective RMSE value. As shown in the figure, the minimum value is obtained at $x= 8$}
  \label{fig:RMSEvsX}
\end{figure}

\section{Analysis and Results}
Table \ref{table:rmse} lists the comparison of the RMSE values for multiple end-to-end models after training them on the Udacity dataset. In addition to the five baseline models listed in Section IV-E, we also include two models of ours: Model F and Model G. Model F is our proposed approach with setting $x = 8$ for each vehicle. Model G sets $x = 10$ time-steps for each vehicle instead of 8 in our model. Since the RMSE values on Udacity dataset were not reported for Model D and Model E in  \cite{Choudhary2017}, we re-implemented those models to compute the RMSE values on Udacity Dataset and reported the results from our implementation in Table \ref{table:rmse} .
 
\begin{table}[b]
\centering
\caption{Comparison to Related Work in terms of RMSE.
{\newline$^{\mathrm{a}}X=8$, $^{\mathrm{b}}X=10$.}}
\begin{center}
 \begin{tabular*}{0.48\textwidth}{@{\extracolsep{\fill} }  c c c c c c c c }
   &  &  &  &  &  &  & \\ 
   Model: &A&B  &C  &D  &E  &F$^{\mathrm{a}}$ & G$^{\mathrm{b}}$ \\
 & \cite{Bojarski2016} & \cite{Du2017} & \cite{Du2017} & \cite{Choudhary2017} & \cite{Choudhary2017} & Ours & Ours
 
 \\ 
 \hline
 \hline
  Training & 0.099 & 0.113 & 0.077 & 0.061 & 0.177 & \textbf{0.034} & 0.044\\
  Validation & 0.098 & 0.112 & 0.077 & 0.083 & 0.149 & \textbf{0.042} & 0.044\\
 \hline
 \hline
 \end{tabular*}
\end{center}
\label{table:rmse}
\end{table}

\begin{table}[b]
\centering
\caption{Comparison to Related Work in terms of MAE.
{\newline$^{\mathrm{a}}X=8$, $^{\mathrm{b}}X=10$.}}
\begin{center}
\begin{tabular*}{0.48\textwidth}{@{\extracolsep{\fill} }  cccccc }
    & A & D & E & F$^{\mathrm{a}}$ & G$^{\mathrm{b}}$\\ 
   Model:& \cite{Bojarski2016} & \cite{Choudhary2017} & \cite{Choudhary2017} & Ours & Ours\\
 \hline
 \hline
  Training & 0.067 & 0.038 & 0.046 & \textbf{0.022} & 0.031\\
  Validation & 0.062 & 0.041 & 0.039 & \textbf{0.033} & 0.036\\
 \hline
 \hline
\end{tabular*}
\end{center}
\label{table:mae}
\end{table}

\par
Table \ref{table:mae} lists the MAE values computed for our implementations of the models A, D, E, F, and G. Models A, B, C, D, and E do not report their individual MAE values in their respective sources. While we re-implemented each of those models in Keras, our implementations of the models B and C yielded higher RMSE values than their reported values even after hyperparameter tuning. Consequently, we did not include the MAE results of our implementations for those two models in Table \ref{table:mae}.  The MAE values for the models A, D and E are obtained after hyperparameter tuning.
\par
We then study the effect of changing the value of $x$ on the performance of our model in terms of RMSE. We train our model at separate $x$ values where $x$ is set to 1, 2, 4, 6, 8, 10, 12, 14, 20 and computed the RMSE value for both the training and validation data respectively at each $x$ value. The results were plotted in Fig. \ref{fig:RMSEvsX}. As shown in the figure, we obtained the lowest RMSE value for both training and validation data at the value when $x = 8$, where $\text{RMSE} = 0.042$ for the validation data. The figure also shows that choosing the appropriate $x$ value is important to receive the best performance from the model. As Fig. \ref{fig:RMSEvsX} shows, the number of the used images in the input affects the performance. Next, we study how changing the $\Delta t$ value affects the performance of our end-to-end system in terms of RMSE value during the testing, once the algorithm is trained at a fixed $\Delta t$.
\par
\begin{figure}[b]
  \centering
 \includegraphics[width=.48\textwidth]{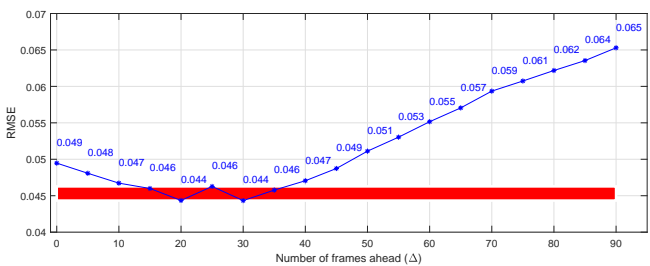}
  \caption{RMSE value vs. size of the number of frames ahead ($\Delta t$) over the validation data. The model is trained at $\Delta t = 30$ and $x = 10$. Between the $\Delta t$ values: 13 and 37 (the red area) the change in RMSE value remains small and the algorithm almost yields the same min value at $\Delta t = 20$ which is different than the training value.}
  \label{fig:RMSE_vsnumber_frames}
\end{figure}

Changing $\Delta t$ corresponds to varying the distance between the two vehicles. For that purpose, we first set $\Delta t= 30$ frames (i.e., 1.5 seconds gap between the vehicles) and trained the algorithm accordingly (where $x = 10$). Once our model was trained and learned the relation between the given input image stacks and the corresponding output value at $\Delta t = 30$, we studied the robustness of the trained system as the distance between two vehicles change during the testing. Fig. \ref{fig:RMSE_vsnumber_frames} demonstrates the results on how the RMSE value changes as we change the distance between the vehicles during the testing. For that, we run the trained model over the entire validation data where the input obtained from the validation data formed at $\Delta t$ values varying between 0 and 95 with increments of 5 frames, and we computed the RMSE value at each of those $\Delta t$ values.

\par
As shown in Fig. \ref{fig:RMSE_vsnumber_frames}, at $\Delta t  = 30$, we have the minimum RMSE value (0.0443) as the training data was also trained by setting $\Delta t  = 30$. However, another (local) minimum value (0.0444), that is almost the same as the value obtained the training $\Delta t$ value, is also obtained at $\Delta t  =20$. Because of those two local minimums, we noticed that the change in error remains small inside the red area as shown in the figure. However, the error does not increase evenly on both sides of the training value ($\Delta t = 30$) as most of the RMSE values within the red area remains on the left side of the training value ($\Delta t = 30$).

\par
\begin{figure}[t]
  \centering
   \includegraphics[width=.48\textwidth]{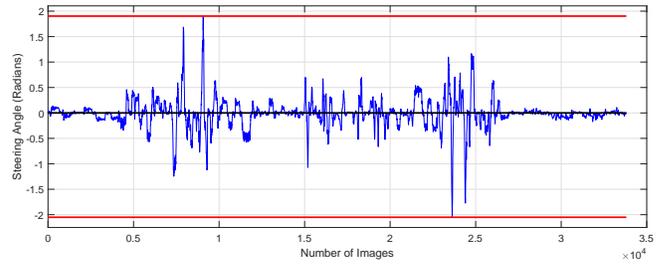}
  \caption{Steering angle (in radians) vs. the index of each image frame in the data sequence is shown for the Udacity Dataset. This data forms the ground-truth for our experiments. The upper and lower red lines highlight the maximum and minimum angle values respectively in the figure.}
  \label{fig:ground_true}
\end{figure}
\begin{figure*}[h]
  \centering
   \includegraphics[width=.90\textwidth]{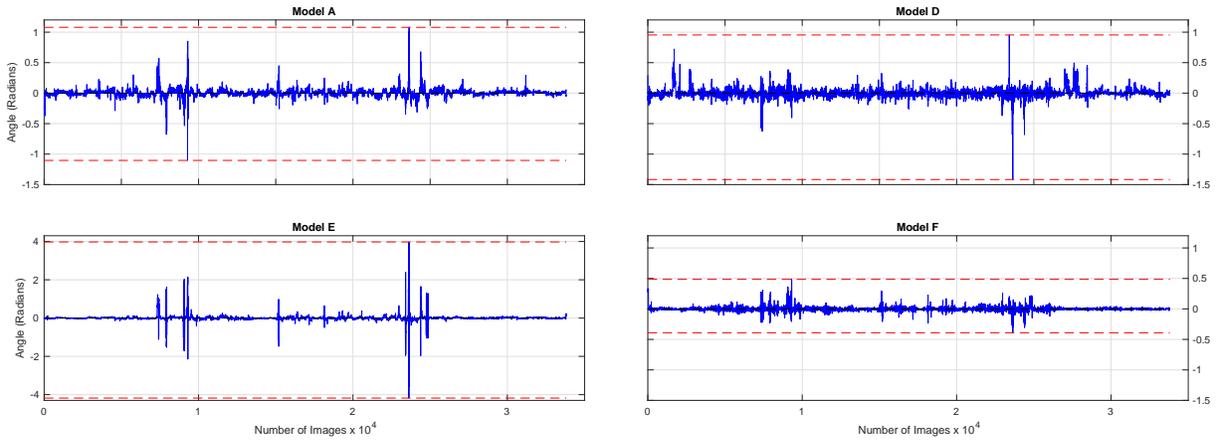}
  \caption{Individual error values (shown in radian) made at each time frame is plotted for four models namely: Model A, Model D, Model E and Model F. The dataset is the Udacity Dataset. Ground-truth is shown in Fig. \ref{fig:ground_true}. The upper and lower red lines highlight the maximum and minimum errors made by each algorithm. The error for each frame (the y axis) for Model A, D and F are plotted in the range of [-1.5, +1,2] and the error for Model E is plotted in the range of [-4.3, +4.3].}
  \label{fig:errors}
  
\end{figure*}

Next, we demonstrate the performance of multiple models over each frame of the entire Udacity dataset in Fig. \ref{fig:errors}. There are total of 33808 images in the dataset. The ground-truth for the figure is shown in Fig. \ref{fig:ground_true} and the difference between the prediction and the ground-truth is given in Fig. \ref{fig:errors} for multiple algorithms. In each plot, the maximum and minimum error values made by each algorithm are highlighted with red lines individually.  In Fig. \ref{fig:errors}, we only demonstrate the results obtained for Model A, Model D, Model E and Model F (ours). The reason for that is the fact that there is no available implementation of Model B and Model C from \cite{Du2017} and our implementations of those models (as they are described in the original paper) did not yield good results to be reported here. Our algorithm (Model F) demonstrated the best performance overall with the lowest RMSE value. Comparing all the red lines in the plots (i.e., comparing all the maximum and minimum error values) suggests that the maximum error made by each algorithm is minimum for our algorithm over the entire dataset.

\section{CONCLUDING REMARKS}
In this paper, we present a new approach by sharing images between cooperative self-driving vehicles to improve the control accuracy of steering angle. Our end-to-end approach uses a deep model using CNN, LSTM and FC layers and our proposed model using shared images yields the lowest RMSE value when compared to the other existing models in the literature.
\par
Unlike previous works that only use local information obtained from a single vehicle, we propose a system where the vehicles communicate with each other and share data. In our experiments, we demonstrate that our proposed end-to-end model with data sharing in cooperative environments yields better performance than the previous approaches that rely on only the data obtained and used on the same vehicle. Our end-to-end model was able to learn and predict accurate steering angles without manual decomposition into road or lane marking detection. 
\par
One potentially strong argument against using image sharing might be that using the geo-spatial information along with the steering angle from the future vehicle and employing the same angle value at that position. Here we argue that using GPS makes the prediction dependent on the location data which, like any other sensor, provides faulty location values in many cases due to various reasons yielding to force algorithms to use wrong image sequence as input. Image sharing over V2V communication helps the model to become resistant to such location-based errors.

\par
We believe that, the reason for skew in Fig. \ref{fig:RMSE_vsnumber_frames} being towards the left, inside the red area is related to the car’s speed. As the car goes faster (which can be considered as increasing the $\Delta t$ value), there is less relevant information in the data that comes from the vehicle ahead (Vehicle 2) potentially yielding higher RMSE values. Furthermore, the distance between each frame also increases as the speed increases making the correlation between the consecutive time frames decrease. In future work, we will also focus on this aspect to analyze the exact reason.
\par
More work and analysis are needed to improve the robustness of the proposed model. While this work relies on the simulated data, we are in the process of collecting real data obtained from actual cars communicating over V2V and will perform more detailed analysis on that larger new data. 

\section*{Acknowledgement}
This work was done as a part of CAP5415 Computer Vision class in Fall 2018 at UCF. We gratefully acknowledge the support of NVIDIA Corporation with the donation of the GPU used for this research.

\balance




\bibliography{refs.bib}{}
\bibliographystyle{unsrt}

\end{document}